\documentclass[conference]{IEEEtran}
\IEEEoverridecommandlockouts
\usepackage{cite}
\usepackage{amsmath,amssymb,amsfonts}
\usepackage{algorithmic}
\usepackage{graphicx}
\usepackage{textcomp}
\usepackage{xcolor}
\def\BibTeX{{\rm B\kern-.05em{\sc i\kern-.025em b}\kern-.08em
    T\kern-.1667em\lower.7ex\hbox{E}\kern-.125emX}}    

\usepackage{amsmath}
\usepackage{amsfonts}
\newcommand{\donotdisplay}[1]{}
\newlength{\lgCase}
\graphicspath{{Figures/}}
\usepackage{siunitx}
\usepackage{multirow}
\usepackage{colortbl,etoolbox,siunitx,xcolor}
\robustify\bfseries
\newcommand{\maxf}[1]{\bfseries #1}
\newcolumntype{P}[1]{>{\centering\arraybackslash}p{#1}}
\usepackage{mathtools}
\def\multiset#1#2{\ensuremath{\left(\kern-.3em\left(\genfrac{}{}{0pt}{}{#1}{#2}\right)\kern-.3em\right)}}
\usepackage{url}

\begin{document}
\title{Serialized Interacting Mixed Membership Stochastic Block Model}

\author{\IEEEauthorblockN{Gaël POUX-M\'EDARD}
\IEEEauthorblockA{\textit{ERIC Lab} \\
\textit{Université de Lyon, 69361}\\
Lyon, France \\
0000-0002-0103-8778}
\and
\IEEEauthorblockN{Julien VELCIN}
\IEEEauthorblockA{\textit{ERIC Lab} \\
\textit{Université de Lyon, 69361}\\
Lyon, France \\
0000-0002-2262-045X}
\and
\IEEEauthorblockN{Sabine LOUDCHER}
\IEEEauthorblockA{\textit{ERIC Lab} \\
\textit{Université de Lyon, 69361}\\
Lyon, France \\
0000-0002-0494-0169}
}


\maketitle

\begin{abstract}
Last years have seen a regain of interest for the use of stochastic block modeling (SBM) in recommender systems. These models are seen as a flexible alternative to tensor decomposition techniques that are able to handle labeled data. Recent works proposed to tackle discrete recommendation problems via SBMs by considering larger contexts as input data and by adding second order interactions between contexts' related elements. In this work, we show that these models are all special cases of a single global framework: the Serialized Interacting Mixed membership Stochastic Block Model (SIMSBM). It allows to model an arbitrarily large context as well as an arbitrarily high order of interactions. 
We demonstrate that SIMSBM generalizes several recent SBM-based baselines.
Besides, we demonstrate that our formulation allows for an increased predictive power on six real-world datasets.
\end{abstract}


\section{Introduction}
Clustering is a core concept of machine learning. Among other applications, it has proven to be especially fit to tackle real-world recommendation problems. A recommendation consists in guessing an output \textit{entity} based on a given context. This context can often be represented as a high dimensional set of input entities. On retail websites for instance, the context could be the ID of a user, the last product she bought, the last visited page, the current month, and so on. Clustering algorithms look for regularities in these datasets to reduce the dimensionality of the input context to its most defining characteristics. Continuing the online retail example, a well design algorithm would spot that a mouse, a keyboard and a computer screen are somehow related buys, and that the next buy is likely to be another computer device. Besides, when given a set of users, subsets of users are likely to have similar interest in a given product if their buying history is similar. We can define such groups of people that share similar behaviors using clustering algorithms; this is called collaborative filtering. One of the most widely used approaches to perform this task relies on tensor decomposition.


\begin{figure}
    \centering
    \includegraphics[width=\columnwidth]{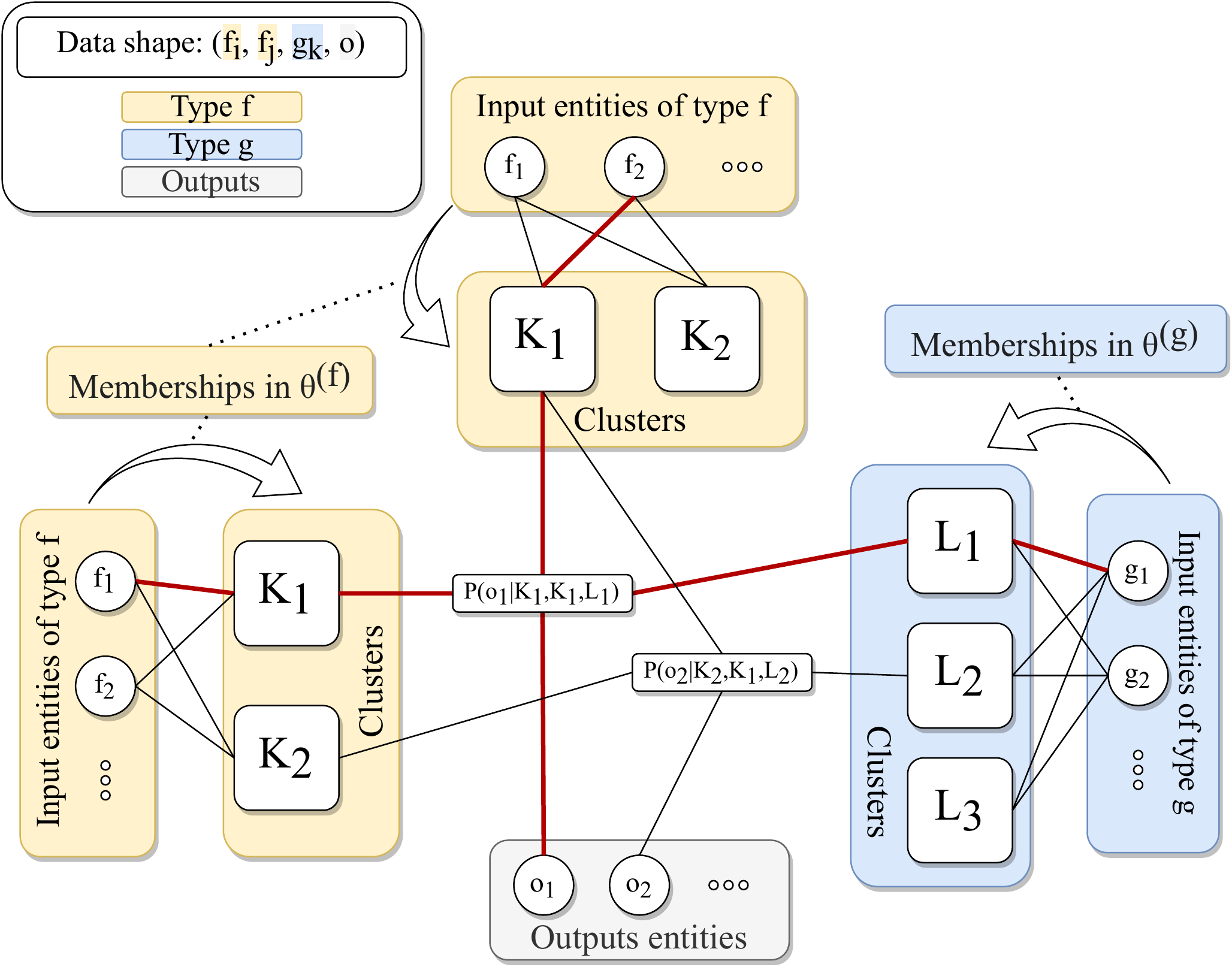}
    \caption{\textbf{Illustration of the SIMSBM} --- For input entities of type $f$ and $g$, where entities of type $f$ interact with each other as pairs and entities of type $g$ do not interact with each other. The membership of an entity $f_i$ of type $f$ to a cluster $K_n$ is encoded into the membership matrix entry $\theta^{(f)}_{f_i,K_n}$. The interaction between clusters is embedded in a multipartite network, whose adjacency matrix is $\mathbf{p}$. A weighted edge between several clusters and one output represents the probability of this output given the context clusters --only two such edges are represented here. \textbf{In red}, we represent the probability of $o_1$ given $f_1$ belonging to $K_1$, $f_2$ belonging to $K_1$ and $g_1$ belonging to $L_1$, which is equal to $\theta_{f_1,K_1}^{(f)}\theta_{f_2,K_1}^{(f)}\theta_{g_1,L_1}^{(g)}P(o_1|K_1,K_1,L_1)$. 
    }
    \label{fig-schema}
\end{figure}

Tensor decomposition approaches provide a variety of efficient mathematical tools for breaking a tensor into a combination of smaller components. One of the most popular tensor decomposition method is Non-negative Matrix Factorization (NMF). Its application to recommender systems has been proposed in 2006 on Simon Funk's blog for an open competition on movie recommendation \cite{Koren2009MFReco}. The underlying idea is to approximate a target 2-dimensional real-valued observations tensor $D \in \mathbb{R}^{I \times J}$ (a matrix in this case) 
as the product of two lower dimensional matrices $W \in \mathbb{R}^{I \times K}$ and $H  \in \mathbb{R}^{K \times J}$ such that $D = WH$.
This approach has seen numerous developments, such as an algorithm allowing its online optimization \cite{Fund2007OnlineNMF1,Cao2007OnlineNMF2}. Thanks to its low computational cost, this method is still today at the core of many real-world large scale recommender systems. However, a major drawback of NMF is that it can only consider two-dimensional data. Several extensions have been proposed to consider n-dimensional data. A straightforward generalization of NMF is the Tensor Factorization, that generalizes NMF in order to infer an n-dimensional matrix $D \in \mathbb{R}^{I \times J \times ...}$ as the product of a core tensor $C \in \mathbb{R}^{K \times L \times ...}$ with n smaller matrices $M_1 \in \mathbb{R}^{I \times K}$, $M_2 \in \mathbb{R}^{J \times L}$, and so on, such that $D = M_1 M_2...M_n C $ \cite{Karatzoglou2010TF}. This approach allows to consider a larger context as input data. Several variants have been proposed based on similar ideas \cite{Hidasi2012Itals,Bhargava2015}.
Another class of tensor decomposition is called Tensor rank or Canonical Polyadic (CP) decomposition. It is at the base of several popular decomposition methods that consider a sum of rank-one matrices instead of a product decomposition. In this case, an n-dimensional tensor is approximated as the sum of rank-one tensors \cite{Harshman1970CP,Carroll1970CP}. Several extensions based on CP have been proposed \cite{Acar2010CP1,Filipovi2015CP2}.

However, decomposition methods are based on linear decomposition of a real-valued tensor $D$, which is unfit to tackle discrete problems. These methods can efficiently infer continuous outputs (the rating of a movie as in \cite{Koren2009MFReco} for instance, or the number of buys of a product) but must be tweaked in order to consider discrete outputs (the next buy on an online retail website for instance). In this case, a possible approach consists in mapping all possible discrete outputs as a continuous variable. This is straightforward as in the case of movie ratings, because the set of possible ratings (1, 2, 3, ...) can be ordered on a continuous scale. However, if one wants to recommend one of several products (mouse, keyboard, computer, ...), the mapping of possible outputs to a continuous value is not trivial.
A way to consider discrete data in decomposition methods to add another dimension to the input tensor, whose size equals the number of possible outputs. Then, the algorithm optimizes the model based on the frequency of each of those items. This trick induces a strong bias and increase the complexity of the algorithm. To answer this problem, recent years have seen a growing interest in the literature of Stochastic Block Modeling (SBM). We will detail this literature after introducing the SIMSBM framework, in order to show how it generalizes most of the state-of-the-art models.

\subsection{Contributions}
In this paper, we propose the Serialized Interacting Mixed membership SBM (abbreviated SIMSBM). The SIMSBM is a global framework that generalizes several state-of-the-art models which tackle the problem of discrete recommendation by a Bayesian network approach. In particular, it generalizes recent works on multipartite graphs inference \cite{Antonia2016AccurateAndScalableRS,Tarres2019TMBM} as well as interaction modeling \cite{Poux2021IMMSBM,Poux2022InterInfSpreadTheWebConf}.
We first introduce the proposed framework in detail and develop an EM optimization procedure. Then, we review previous works on Stochastic Block Models that are used in collaborative filtering, and detail for each one how to recover it as a special case of our framework. Experiments are then conducted on 6 real-world datasets and compared to standard baselines of the literature. We show our formulation allows to obtain better recommendations than existing methods either by adding layers to the modeled multipartite graph or by adding higher-order interactions terms in the modeling.

\section{The SIMSBM framework}
\label{model}
The goal of the SIMSBM is to recommend an \textit{output entity}, that is one of $O$ possible output entities, given a context. To do so, it considers data in the form of a multipartite network. The network's nodes are the context elements (called \textit{input entities} $f_n$, each of which is one of $F_n$ possible input entities). One hyper-edge between the $N$ input entities (or nodes) of a context represents the probability of a given output entity $o$, that is ${P(o \in O \vert f_1 \in F_1, ..., f_N \in F_N)}$. The SIMSBM clusters input entities (or nodes) together and infers edges between these clusters to form a smaller multipartite network. This projection is illustrated Fig.~\ref{fig-schema}. The notations used throughout this section are recalled Table.~\ref{table-not}.

Formally, input data of the SIMSBM is noted ${R^{\circ}}$, and is a collection of tuples ${(f_1, ..., f_N, o)}$. The vector ${\vec{f}=(f_1, ..., f_N)}$ represents a context, whose entries $f_n$ are the input entities that can be one of $F_n$ possible input entities; outputs are designated by $o$. Each input entity is represented as a node in a layer of the multipartite network (circles in Fig.~\ref{fig-schema}). The number of layers of the multipartite graph is then $N = \vert \vec{f} \vert$; each layer $n$ comprises $F_n$ nodes. For each node in each layer, the SIMSBM infers a vector ${\vec{\theta_i} \in \mathbb{R}^{K}}$ that represents its probability to belong to each of $K$ possible clusters (i.e. associate each circle to a distribution over the squares in Fig.~\ref{fig-schema}). 

Input entities are of a given \textit{type}; entities of the same type carry the same semantic meaning, and are drawn from the same set of possible values. We note $a(f)$ the function that associates an input entity $f$ to its given type, and $K_{a(f_n)}$ the number of available clusters for this type. Entities of same type can interact with each other but are forced to share the same cluster membership matrix. For instance, consider a user rating a movie according to the pair of actors starring in it: the context vector takes the shape $(user, actor 1, actor 2)$. In this example, one entity is of type ``user'', and the two other are of type ``actor'', and each of the three entities is embedded in its own layer of a multipartite graph. When creating clusters, the SIMSBM enforces that the two layers of type ``actor'' share the same membership matrix, while the layer accounting for the type ``user'' has its own membership matrix. This structure is exactly the same as the one depicted in Fig.~\ref{fig-schema}. This is needed in order to get results that are permutation independent, meaning in our example that $P(o \vert (user, actor 1, actor 2)) = P(o \vert (user, actor 2, actor 1))$. Our formulation as a multipartite network allows to consider higher order interactions, with contrast to \cite{Poux2021IMMSBM} which only considers pair interactions.

The membership of an entity must sum to 1 over all the $K_{a(f_n)}$ available clusters, hence the following constraint:
\begin{equation}
\label{eq:normTheta}
    \sum_k^{K_{a(f_n)}} \theta_{f_n,k}^{(a(f_n))} = 1 \,\,\forall f_n
\end{equation}
Once the nodes membership is known, the SIMSBM infers the clusters multipartite network, whose weighted hyper-edges stand for the probability of an output $o$ given a combination of clusters. 
The hyper-edge corresponding to the clustered context ${\vec{k}=(k_1, ..., k_N)}$ associated to the output $o$ is written ${p_{\vec{k}}(o) \in \mathbb{R}^{K_{a(f_1)} \times ... \times K_{a(f_N)} \times O}}$.
Note that the clustered context $\vec{k}$ can take any value among all the possible permutations ${\vec{K} = \{K_1, ..., K_N\}_{K_1=1, ..., K_{a(f_1)}; ...; K_N=1,...,K_{a(f_N)}}}$.
As we want SIMSBM to infer a distribution over possible outputs in a given context, the edges of the multipartite graph are related by the following constraint:
\begin{equation}
    \sum_o p_{\vec{k}}(o) = 1 \,\,\,\, \forall \vec{k} \in \vec{K}
\end{equation}
Finally, the probability of an output $o$ given a context of input entities $\vec{f}$ can be written as:
\begin{equation}
    \label{eq-probfinale}
    P(o \vert\vec{f}) = \sum_{\vec{k} \in \vec{K}} p_{\vec{k}}(o) \prod_{n \in N} \theta_{f_n,k_n}^{(a(f_n))}
\end{equation}
From Eq.\ref{eq-probfinale}, we can define the log-likelihood of the model as:
\begin{equation}
\label{eq:likelihood}
   \ell= \sum_{(\vec{f},o) \in R^{\circ}} \log \left( \sum_{\vec{k} \in \vec{K}} p_{\vec{k}}(o) \prod_{n \in N} \theta_{f_n,k_n}^{(a(f_n))} \right)
\end{equation}

\begin{table}
    \caption{Notations}
    \label{table-not}
	\centering
	\begin{tabular}{|l|l|}
	$f_n$ & An input entity, can take any value in $F_n$\\
	$F_n$ & Set of possible input entities for layer $n$\\
	$\vec{f}$ & Context vector $(f_1, ..., f_N)$\\
	$o$ & Output entity, can take any value in $O$ \\
	$N$ & Number of input layers $\vert \vec{f} \vert$ \\
	$R^{\circ}$ & Data, a list of (N+1)-plets $(f_1, ..., f_N, o)$\\
	$a(f_n)$ & type of entity $f_n$\\
	$K_{a(f_n)}$ & Number of available clusters for type $a(f_n)$\\
	$\theta^{(a(f_n))}$ & Membership matrix for entities of type $a(f_n)$ \\
	$\mathbf{p}(o)$ & Clusters' multipartite network for output $o$\\
	$\vec{K}$ & Every possible clusters permutation $\{(k_1,..., k_N)\}_{k_1;...;k_N}$\\
	$\vec{k}$ & One permutation of clusters indices $(k_1,..., k_N) \in \vec{K}$ \\
	$c_v(x)$ & Count of element $x$ in vector $\vec{v}$\\
	$C_{f_n}$ & Total count of $f_n$ in $R^{\circ}$\\
	\end{tabular}
\end{table}

\subsection{Inference}
In this section, we derive an Expectation-Maximization algorithm for inferring the model's parameters $\mathbf{p}, \mathbf{\theta}$. Such algorithm guarantees the convergence towards a local maximum of the likelihood function \cite{Neal1998ConvergenceEM}.

\subsubsection{E-step (short derivation)}
Using Jensen's inequality, we can rewrite Eq.\ref{eq:likelihood} as:
\begin{align}
    \ell 
    &= \sum_{(\vec{f},o) \in R^{\circ}} \log \left( \sum_{\vec{k} \in \vec{K}} p_{\vec{k}}(o) \prod_{n \in N} \theta_{f_n,k_n}^{(a(f_n))} \right) \\
    &\geq \sum_{(\vec{f},o) \in R^{\circ}} \sum_{\vec{k} \in \vec{K}} \omega_{\vec{f},o}(\vec{k}) \cdot \log  \left( \frac{p_{\vec{k}}(o) \prod_{n \in N} \theta_{f_n,k_n}^{(a(f_n))}}{\omega_{\vec{f},o}(\vec{k})} \right) \notag
\end{align}
The inequality holds as an equality when:
\begin{equation}
\label{eq:expectation}
    \omega_{\vec{f},o}(\vec{k}) = \frac{p_{\vec{k}}(o) \prod_{n \in N} \theta_{f_n,k_n}^{(a(f_n))}}{\sum_{\vec{k'} \in \vec{K}} p_{\vec{k'}}(o) \prod_{n \in N} \theta_{f_n,k'_n}^{(a(f))}}
\end{equation}
Eq.\ref{eq:expectation} constitutes the expectation step of the EM algorithm. This derivation is intended as a fast way of deriving the correct result. An alternative method that is more explicit about the underlying concepts used in the derivation is described in the follow-up subsection; the final expression for $\omega_{\vec{f},o}(\vec{k})$ however is identical. 

\subsubsection{E-step (detailed derivation)}
The derivation presented in this section follows a well-known general derivation of the EM algorithm, which can be found in C.M. Bishop's \textit{Pattern Recognition and Machine Learning}-p.450 for instance.


We recall that the total log-likelihood (Eq.~\ref{eq:likelihood}) is simply the sum of each obervation's log-likelihood: $\log P(R^{\circ} \vert \theta, p) = \sum_{(\vec{f}, o) \in R^{\circ}}\log P(\vec{f}, o \vert \theta, p)$. Without loss of generality, we focus here on a single observation for clarity. The expression of the SIMSBM log-likelihood for one entry entry of the dataset reads:
\begin{align}
\label{eq-L}
    \log P(\vec{f}, o \vert \theta, p)
    &= \log \sum_{\vec{k} \in \vec{K}} P(\vec{f}, o, \vec{k} \vert \theta, p) \\
    &= \log \left( \sum_{\vec{k} \in \vec{K}} p_{\vec{k}}(o) \prod_{n \in N} \theta_{f_n,k_n}^{(a(f_n))} \right) \notag
\end{align}

We assume a distribution $Q(\vec{k})$ on the latent variables (accounting for cluster allocation) associated to one observation in the dataset $R^{\circ}$; this distribution is yet to be defined. Because $\vec{k}$ takes values in $\vec{K}$, we have $\sum_{\vec{k} \in \vec{K}} Q(\vec{k}) = 1$. Given this normalization condition, we can decompose Eq.\ref{eq-L} for any distribution $Q(\vec{k})$ as:

\begin{align}
    \label{eq-decomp}
    \log &P(\vec{f}, o \vert \theta, p) = \underbrace{\log P(\vec{f}, o, \vec{k} \vert \theta, p) - \log P(\vec{k} \vert \vec{f}, o, \theta, p)}_{\text{Does not depend on $\vec{k}$}} \notag \\
    =\ \ \ \  &\sum_{\vec{k} \in \vec{K}} Q(\vec{k}) \left( \log P(\vec{f}, o, \vec{k} \vert \theta, p) - \log P(\vec{k} \vert \vec{f}, o, \theta, p) \right) \notag \\
    =\ \ \ \  &\sum_{\vec{k} \in \vec{K}} Q(\vec{k}) \log P(\vec{f}, o, \vec{k} \vert \theta, p) \notag \\
    - &\sum_{\vec{k} \in \vec{K}} Q(\vec{k}) \log P(\vec{k} \vert \vec{f}, o, \theta, p) \notag \\
    =\ \ \ \  &\sum_{\vec{k} \in \vec{K}} Q(\vec{k}) \log \frac{P(\vec{f}, o, \vec{k} \vert \theta, p)}{Q(\vec{k})}  \notag \\
    - &\sum_{\vec{k} \in \vec{K}} Q(\vec{k}) \log \frac{P(\vec{k} \vert \vec{f}, o, \theta, p)}{Q(\vec{k})}
\end{align}

We note that the term in the last line of Eq.\ref{eq-decomp}, $- \sum_{\vec{k} \in \vec{K}} Q(\vec{k}) \log \frac{P(\vec{k} \vert \vec{f}, o, \theta, p)}{Q(\vec{k})}$, is the Kullback-Leibler (KL) divergence between $P$ and $Q$, noted $KL(P \vert \vert Q)$. The KL divergence obeys $KL(P \vert \vert Q) \geq 0$, and is null iif $P$ equals $Q$. Therefore, the term in the before-last line of Eq.\ref{eq-decomp}, $\sum_{\vec{k} \in \vec{K}} Q(\vec{k}) \log \frac{P(\vec{f}, o, \vec{k} \vert \theta, p)}{Q(\vec{k})}$, is interpreted as a lower bound on the log-likelihood $\log P(\vec{f}, o \vert \theta, p)$. 

The aim of the E-step is to find the expression of $Q(\vec{k})$ that maximizes the lower bound of the log-likelihood with respect to the latent variables $\vec{k}$. Given that the log-likelihood does not depend on $Q(\vec{k})$ and $KL(P \vert \vert Q) \geq 0$, the lower-bound is maximized when $KL(P \vert \vert Q) = 0$, which occurs when $Q(\vec{k}) = P(\vec{k} \vert \vec{f}, o, \theta, p)$. In this case, the lower-bound on the log-likelihood equals the likelihood itself and thus reaches a maximum with respect to the latent variables $\vec{k}$ for fixed parameters $\theta$ and $p$.

Given the definition of the SIMSBM, the derivation of $P(\vec{k} \vert \vec{f}, o, \theta, p)$ is straightforward (see Eq.\ref{eq-probfinale}). The probability of one combination of clusters $\vec{k}$ among $\vec{K}$ possible combinations given an input features vector and an output $o$ is proportional to $p_{\vec{k}}(o) \prod_{n \in N} \theta_{f_n,k_n}^{(a(f_n))}$. Therefore:

\begin{align}
    P(\vec{k} \vert \vec{f}, o, \theta, p) := \omega_{\vec{f}, o}(\vec{k}) = \frac{p_{\vec{k}}(o) \prod_{n \in N} \theta_{f_n,k_n}^{(a(f_n))}}{\sum_{\vec{k'} \in \vec{K}} p_{\vec{k'}}(o) \prod_{n \in N} \theta_{f_n,k'_n}^{(a(f_n))}}
\end{align}
which we denote as $\omega_{\vec{f}, o}(\vec{k})$ for simplicity of notation. 

The EM is a 2-step iterative algorithm. The expression of $P(\vec{k} \vert \vec{f}, o, \theta, p)$ is computed first during the E-step using the parameters $\theta$ and $p$ from the previous iteration, that we note $\theta^{(old)}$ and $p^{(old)}$. 
Once this expression is found, the log-likelihood maximized with respect to the latent variables $\vec{k}$ can be expressed as:
\begin{align}
\label{eq-maxLik}
    &\log P(\vec{f}, o \vert \theta, p) \\
    &= \sum_{\vec{k} \in \vec{K}} P(\vec{k} \vert \vec{f}, o, \theta^{(old)}, p^{(old)}) \log \frac{P(\vec{f}, o, \vec{k} \vert \theta, p)}{P(\vec{k} \vert \vec{f}, o, \theta^{(old)}, p^{(old)})} \notag
\end{align}
The maximization step follows by maximizing Eq.\ref{eq:likelihood} (which is the sum of Eq.~\ref{eq-maxLik} over all observations) with respect to the parameters $\theta$ and $p$, which do not appear in $P(\vec{k} \vert \vec{f}, o, \theta^{(old)}, p^{(old)})$. This derivation is detailed in the follow-up section.

\subsubsection{M-step}
We take back Eq.\ref{eq:likelihood} and add Lagrangian multipliers $\phi$ to account for the constraints on $\mathbf{\theta}$. We maximize of the resulting constrained likelihood $\ell_c$ according to each latent variable:
\begin{align}
\label{eq:m-theta}
    &\frac{\partial \ell_c}{\partial \theta_{mn}^{(a(m))}} = \frac{\partial}{\partial \theta_{mn}^{(a(m))}} \left[ \ell - \sum_i \phi_i^{(a(i))} \left( \sum_k \theta_{ik}^{(n)} - 1 \right) \right] \notag\\
    &\Leftrightarrow \ \ \phi_{m}^{(a(m))} = 
     \sum_{(\vec{f},o) \in \partial m} \sum_{\vec{k} \in \vec{K}} \frac{ c_k(n) \omega_{\vec{f},o}(\vec{k})}{\theta_{mn}^{(a(m))}}  \notag \\
    &\Leftrightarrow \ \ \theta_{mn}^{(a(m))} \phi_{m}^{(a(m))} = 
     \sum_{(\vec{f},o) \in \partial m} \sum_{\vec{k} \in \vec{K}} c_k(n) \omega_{\vec{f},o}(\vec{k})
\end{align}
The term $c_k(n)$ arises because of the non-linearity induced by the interaction between input entities of the same type.
Let $i_m$ be the indices where entity $m$ appears in the input vector $\vec{f}$. The corresponding entries of the permutation vector $\vec{k}$ are noted $\vec{k_{i_m}}$. Then, ${c_k(n)=\vert \{ 1 \vert \vec{k}_{i}=n \}_{i \in i_m} \vert}$ is the count of $n$ in $\vec{k}_{i_m}$.
When $n$ appears $c_k(n)$ times in a permutation comprising $\vec{k_{i_m}}$, so will a term $\log \theta_{nm}^{c_k(n)}$, whose derivative is $\frac{c_k(n)}{\theta_{nm}}$, hence this term arising. 
Note that $c_k(n)=0$ nullifies the contribution of permutations $\vec{k}$ where $n$ does not appear in $\vec{k_{i_m}}$. Therefore we can restrict the sum over $\vec{k}$ in Eq.\ref{eq:m-theta} to the set ${\partial n = \{ \vec{k} \vert \vec{k}\in \vec{K}, n \in \vec{k}_{i_m} \}}$. We also defined the set ${\partial m = \{ (\vec{f}, o) \vert (\vec{f}, o) \in R^{\circ}, m \in \vec{f}_{i_m} \}}$. 

Using Eq.\ref{eq:normTheta} and Eq.\ref{eq:m-theta}, we compute $\phi_m^{(a(m))}$:
\begin{align}
\label{eq:phi}
    \sum_n^{K_{a(m)}} \phi_{m}^{(a(m))}& \theta_{mn}^{(a(m))}  = \sum_{(\vec{f},o) \in \partial m} \sum_n^{K_{a(m)}} \sum_{\vec{k} \in \vec{K}}  c_k(n) \omega_{\vec{f},o}(\vec{k})  \notag \\
    = \phi_{m}^{(a(m))} &= \sum_{(\vec{f},o) \in \partial m}  \underbrace{\sum_{\vec{k} \in \vec{K}} \omega_{\vec{f}, o}(\vec{k})}_{=1 \text{ (Eq.~\ref{eq:expectation})}} \underbrace{\sum_n^{K_{a(n)}} c_k(n)}_{=c_f(m)}  \notag \\
    &= \sum_{(\vec{f},o) \in \partial m}  c_f(m) = C_m
\end{align}
When summing over $n$, $c_k(n)$ successively counts the number of times each $n$ appears in $\vec{k_{i_m}}$, which equals the length of $\vec{k_{i_m}}$. Therefore $\sum_i n_i = \vert \vec{k_{i_m}} \vert = c_f(m)$ is the number of times input entity $m$ appears in the entry $(\vec{f},o)$ considered, which does not depend on $\vec{k}$. $C_m$ is the total count of $m$ in the dataset. Note that this differs from the derivation proposed in \cite{Tarres2019TMBM}, where nonlinear terms are not accounted for. 

The derivation of the maximization equation for $\mathbf{p}$ is very similar. Let $\partial s = \{ (\vec{f}, o) \vert (\vec{f}, o) \in R^{\circ}, o = s \}$. We solve:
\begin{align}
    \label{eq:max-p}
    &\frac{\partial \ell_c}{\partial p_{\vec{r}}(s)} = \frac{\partial}{\partial p_{\vec{r}}(s)} \left[ \ell - \sum_{\vec{k}} \psi_{\vec{k}} \left( \sum_o p_{\vec{k},o} - 1 \right) \right]=0 \notag\\
    &\Leftrightarrow \psi_{\vec{r}} = \sum_{(\vec{f}, o) \in \partial s}  \frac{\omega_{\vec{f},o}(\vec{r})}{p_{\vec{r}}(s)} \notag\\
    &\Leftrightarrow \sum_{n}\psi_{\vec{r}} p_{\vec{r}}(s) = \psi_{\vec{r}} = \sum_{(\vec{f}, o) \in R^{\circ}}  \omega_{\vec{f},o}(\vec{r})
\end{align}
Finally, combining Eq.\ref{eq:m-theta} with Eq.\ref{eq:phi}, and the two last lines of Eq.\ref{eq:max-p}, the maximization equations are:
\begin{equation}
    \begin{cases}
        \label{eq:maxTheta}
        \theta_{mn}^{(a(m))} = \frac{\sum_{(\vec{f},o) \in \partial m} \sum_{\vec{k} \in \partial n} c_k(n) \omega_{\vec{f},o}(\vec{k})}{C_m}\\
        p_{\vec{r}}(s) = \frac{\sum_{(\vec{f}, o) \in \partial s} \omega_{\vec{f},o}(\vec{r})}{\sum_{(\vec{f}, o) \in R^{\circ}} \omega_{\vec{f},o}(\vec{r})}
    \end{cases}
\end{equation}
From Eq.\ref{eq:maxTheta} we can show that for a given number of clusters for each type $(K_{a(f_1)}, ..., K_{a(f_N)})$, one iteration of the EM algorithm runs in $\mathcal{O}(\vert R^{\circ} \vert)$.

We must define a nomenclature to refer to each special case of the SIMSBM --what input entity types are considered, and how many interactions for each type. We use the notation SIMSBM(number interactions type 1, number interactions type 2, ...). For instance, SIMSBM(2,3) represents a case where the SIMSBM considers two types of input entities, with the first one interacting as pairs with other entities of same type, and the second one interacting as triples with entities of the same type. The corresponding data has a shape $(f_1, f_2, g_1, g_2, g_3, o)$ where $f$ and $g$ are the two considered types.

\section{Background on Stochastic Block Models}
\subsection{Existing works}
As stated in the introduction, recent years saw a growing interest for Stochastic Block Models (SBM) to tackle collaborative filtering problems in recommender systems \cite{Antonia2016AccurateAndScalableRS,Tarres2019TMBM,Poux2021IMMSBM}. These models first cluster input entities together, and then analyze how these clusters relate to each other. Each input entity can be associated either to one cluster only (single-membership SBM) \cite{Holland1983SBM,Guimera2012HumanPrefSBM,Funka2019ReviewSBM} or to a distribution over available clusters (Mixed Membership SBM, or MMSBM) \cite{Airoldi2008MMSBM}. While the single-membership SBM has been successfully applied to a range of problems \cite{Guimera2013DrugdrugSBM,Guimera2012HumanPrefSBM,CoboLopez2018SocialDilemma,Funka2019ReviewSBM}, inference is done using greedy algorithms, typically simulated annealing, making it unfit for large scale real-world applications \cite{Antonia2016AccurateAndScalableRS}. 

The Mixed-Membership SBM (MMSBM) is a major extension of Single-Membership SBM that has been proposed in the seminal work \cite{Airoldi2008MMSBM}. 
In the frame of collaborative filtering for recommender systems, \cite{Antonia2016AccurateAndScalableRS} proposed a bipartite network extension. This model has later been extended to consider triples of input entities instead of pairs \cite{Tarres2019TMBM}. It assumes that all the entities in a given triplet are linked together by a given relation. This boils down to assuming data can be represented in the form of a tripartite network instead of a bipartite network.
Another extension of \cite{Airoldi2008MMSBM} proposes to consider the case of input entities of the same type, modeled as a bipartite graph \cite{Poux2021IMMSBM}. This is relevant when trying to guess an output given a pair of input entities of same type -- that is when entities carry the same semantic meaning. When considering input entities of the same type, one must consider symmetric clustering; the probability of an output based on the entities pair $(A,B)$ should not differ from the probability of an output based on $(B,A)$. The authors solve the problem by clustering both entities using a same membership matrix, whose components then interact with each other. This differs from other recent works on interaction modeling that do not consider clustering \cite{Christakopoulou2014HOSLIM,Steck2021Interactions} or the non-linearity induced by symmetric interactions \cite{Myers2012CoC,Tarres2019TMBM}.

\subsection{SIMSBM generalizes several state-of-the-art models}

The formulation of SIMSBM allows to recover several state-of-the-art works. Each of these previous models was introduced as different and novel in their respective publications, whereas they could be presented as simple iterations of SIMSBM instead. Building on this generalization, SIMSBM provides a degree of modeling flexibility that goes beyond the existing literature --modeling arbitrarily large context sizes and interaction order. 

Now, we briefly show how our formulation allows to recover several state-of-the-art models. Throughout this section, we denote input entities of different types by different letters (e.g. $f_1$ is not of the same type as $g_1$), and the model's output as $o$. The set of corresponding membership matrices for each type is noted as $\Theta = \{ \theta^{(f)}, \theta^{(g)}, ... \}$ and one edge of the multipartite clusters-interaction tensor is noted $(p_{k(f_1), k(f_2), ...}(o))$ where $k(f_i)$ is one of the possible cluster indices for an input entity of type $f$.

\subsubsection{MMSBM \cite{Airoldi2008MMSBM}}
The historical MMSBM has been proposed in \cite{Airoldi2008MMSBM}, and is at the base of most models discussed in this section. MMSBM takes pairs $(f_1, o)$ as input data. We can recover this model with our framework by setting $\Theta = \{ \theta^{(f)} \}$. The multipartite network then becomes ``unipartite'', that is a simple one-layer clustering of entities. The probability of an output is defined by entities' cluster membership only. The tensor $p$ then takes the shape $p_{f_1}(o)$. Using the SIMSBM notation, this correspond to SIMSBM(1).

\subsubsection{Bi-MMSBM \cite{Antonia2016AccurateAndScalableRS}}
The Bi-MMSBM has first been proposed in \cite{Antonia2016AccurateAndScalableRS}, and has since been applied on several occasions \cite{Tarres2019TMBM,Poux2021MMSBMMrBanks}. In this modeling, data is made of triplets $(f_1, g_1, o)$. Each entity is associated a node on a side of a bipartite graph ($f_i$'s on one side, $g_i$'s on the other) and edges represent the probability of an output $o$.
We recover the Bi-MMSBM with our model by setting $\Theta = \{ \theta^{(f)}, \theta^{(g)} \}$ and the bipartite clusters network tensor $(p_{k(f_1), k(g_1)}(o))$. This correspond to SIMSBM(1,1).

\subsubsection{T-MBM \cite{Tarres2019TMBM}}
The T-MBM is a model proposed in \cite{Tarres2019TMBM} that goes a step further than \cite{Antonia2016AccurateAndScalableRS} by adding a layer to the bipartite network used to model quadruplet data $(f_1, f_2, g_1, o)$. This model aims at modeling interactions between entities of same type as in \cite{Poux2021IMMSBM} by clustering $f_1$ and $f_2$ using a same membership matrix, but does not account for nonlinear terms. 
We recover the T-MBM modeling by setting $\Theta = \{ \theta^{(f)}, \theta^{(g)} \}$ and $(p_{k(f_1), k(f_2), k(g_1)}(o))$. Our formulation allows to go further by adding an arbitrary number of layers to the multipartite networks proposed in \cite{Antonia2016AccurateAndScalableRS,Tarres2019TMBM}. This correspond to SIMSBM(2,1).

\subsubsection{IMMSBM \cite{Poux2021IMMSBM}}
The IMMSBM proposed in \cite{Poux2021IMMSBM} models interactions between entities of the same type to predict an output. The data takes the form $(f_1, f_2, o)$. Each input entity is still associated to one node on either side of a bipartite graph, except that here the membership matrix is shared between the two layers. The links between each pair of clusters represent the probability of an output $o$.
We recover the IMMSBM with our model by setting $\Theta = \{ \theta^{(f)} \}$ and the bipartite clusters network tensor $(p_{k(f_1), k(f_2)}(o))$. Importantly, our formulation allows to consider interactions between $n$ input entities instead of simply pair interactions. This correspond to SIMSBM(2).

\begin{table*}
    \setlength{\lgCase}{2.25cm}
    \caption{Datasets considered. The number of discrete values each input or output entity type can can take is given between parenthesis.}
    \label{table-DS}
	\centering
	\noindent\makebox[\textwidth]{\resizebox{\textwidth}{!}{
	\begin{tabular}{|P{0.7\lgCase}|P{3.05\lgCase}|P{0.85\lgCase}|P{0.90\lgCase}|P{0.3\lgCase}|P{0.6\lgCase}|}
    \cline{1-6}
    & Type of the input entities & \# interactions & Type outputs & $\vert R^{\circ} \vert$ & \# clusters\\
    \cline{1-6}
    
    MrBanks 1 & \{Player (280), Situation (7), Gender (2), Age (6)\} & Situations: 3 & User guess (2) & 16k & \{5,5,3,3\} \\
    MrBanks 2 & \{Player (280), Situation (7)\} & Situations: 3 & User guess (2) & 16k & \{5,5\} \\
    Spotify & \{Artists (143)\} & Songs: 3 & Artist (740) & 50k & \{20\} \\
    PubMed & \{Symptoms (13)\} & Symptoms: 3 & Disease (280) & 2M & \{20\} \\
    Imdb 1 & \{User (2502), Casting (809)\} & Casting: 2 & Rating (10) & 1M & \{10,8\} \\
    Imdb 2 & \{User (2502), Director (255), Casting (809)\} & None & Rating (10) & 700k & \{10,10,10\} \\

    \cline{1-6}
	\end{tabular}
	}}
\end{table*}
\subsubsection{Indirect generalizations}
We did not detail the generalization of other families of block models because our algorithm does not readily fits these cases. However, it is worth mentioning that MMSBM has been developed as an alternative to Single Membership SBM \cite{Yuchung1987} that allows more flexibility \cite{Airoldi2008MMSBM}. Our model reduces to most existing SBM by modifying the definition of the entries of $\theta^{(n)}$. In the Single Membership SBM, Eq.\ref{eq:normTheta} reads $\theta_{f_n, k}^{(n)}=\delta_{k, x}$ where $x$ is one of the $K_n$ possible values for $k$ and $\delta$ is the Kronecker's delta. This means the membership vector of each input entity equals 1 for one cluster only, and 0 anywhere else. Therefore, the optimization process is not the same as we described. In practice optimization is done with greedy algorithms such as the simulated annealing \cite{CoboLopez2018SocialDilemma,Poux2021MMSBMMrBanks}.

It has also has been shown in \cite{Antonia2016AccurateAndScalableRS}-Eq.7 that the Bi-MMSBM model generalizes matrix factorization. Therefore, it follows that SIMSBM also generalizes it. The underlying idea is to remove the multipartite network tensor $p$ and define clusters that are shared by both sides of the bipartite network. This way, clusters do not interact with each other because they are not embedded into a multipartite network; input entities on one side of the bipartite network belonging to one cluster are solely linked to entities on the other side belonging to this same cluster.

\section{Experiments}
\label{Experiments}
\subsection{Range of application}
As shown in the previous section, our formulation generalizes several existing models from the state-of-the-art. Therefore, it is readily applicable to any of the datasets considered in these works. This includes recommendation datasets (movies \cite{Antonia2016AccurateAndScalableRS}, songs \cite{Poux2021IMMSBM}), medical datasets (symptoms-disease networks \cite{Poux2021IMMSBM}, drug interaction networks \cite{Tarres2019TMBM,Guimera2013DrugdrugSBM}) and social behavior datasets (social dilemmas \cite{CoboLopez2018SocialDilemma,Poux2021MMSBMMrBanks}, e-mail networks \cite{Tarres2019TMBM,Godoy2016EmailNetwork}). In general, it applies to datasets where there is a given number of input entities leading to a set of possible outputs. In this section, we propose to illustrate an application of our model on 6 different datasets.

\begin{table*}
    \caption{Results for every dataset presented. The letters in superscript represent the model SIMSBM generalizes in this particular configuration (MMSBM \cite{Airoldi2008MMSBM}=$^{\text{a}}$; Bi-MMSBM \cite{Antonia2016AccurateAndScalableRS}=$^{\text{b}}$; IMMSBM \cite{Poux2021IMMSBM}=$^{\text{c}}$; T-MBM \cite{Tarres2019TMBM}=$^{\text{d}}$). The standard error on the last digits over all 100 runs is indicated between parenthesis. Overall, we see that our formulation allows to improve results on every dataset.}
    \label{table-res}
	\centering
	\noindent\makebox[\textwidth]{\resizebox{\textwidth}{!}{
	\begin{tabular}{|l|l|l|S|S|S|S|S|S|S}

		\cline{1-9}
		& & & \text{F1} & \text{P@1} & \text{AUCROC} & \text{AUCPR} & \text{RankAvgPrec} & \text{CovErrNorm} \\ 

		\cline{1-9}
		\multirow{8}{*}{\rotatebox[origin=c]{90}{\footnotesize \text{\textbf{MrBanks 1}}}}

		& \multirow{8}{*}{\rotatebox[origin=c]{90}{\scriptsize \text{\textbf{Ply, Sit (3), Gen, Age}}}}
		& SIMSBM(1,1,1,1) &  \num{ 0.7124 +- 0.0002 } &  \num{ 0.6549 +- 0.0003 } &  \num{ 0.7071 +- 0.0002 } &  \num{ 0.7141 +- 0.0003 } &  \num{ 0.8274 +- 0.0001 } &  \num{ 0.1726 +- 0.0001 } \\ 
		& & SIMSBM(1,2,1,1) &  \num{ 0.7107 +- 0.0002 } &  \num{ 0.6696 +- 0.0005 } &  \num{ 0.7120 +- 0.0004 } &  \num{ 0.7158 +- 0.0005 } &  \num{ 0.8348 +- 0.0003 } &  \num{ 0.1652 +- 0.0003 } \\ 
		& & SIMSBM(1,3,1,1) &  \maxf{ \num{ 0.7348 +- 0.0002 } } & \maxf{ \num{ 0.7172 +- 0.0005 } } & \maxf{ \num{ 0.7610 +- 0.0004 } } & \maxf{ \num{ 0.7646 +- 0.0004 } } & \maxf{ \num{ 0.8586 +- 0.0003 } } & \maxf{ \num{ 0.1414 +- 0.0003 } } \\ 
		& & TF &  \num{ 0.6795 } &  \num{ 0.6037 } &  \num{ 0.4702 } &  \num{ 0.4967 } &  \num{ 0.8019 } &  \num{ 0.1981 } \\ 
		& & NMF &  \num{ 0.7178 } &  \num{ 0.6976 } &  \num{ 0.7232 } &  \num{ 0.7182 } &  \num{ 0.8409 } &  \num{ 0.1591 } \\ 
		& & KNN &  \num{ 0.7023 } &  \num{ 0.6648 } &  \num{ 0.6859 } &  \num{ 0.6623 } &  \num{ 0.8324 } &  \num{ 0.1676 } \\ 
		& & NB &  \num{ 0.6867 } &  \num{ 0.6382 } &  \num{ 0.6323 } &  \num{ 0.6250 } &  \num{ 0.8191 } &  \num{ 0.1809 } \\ 
		& & BL &  \num{ 0.6795 } &  \num{ 0.6037 } &  \num{ 0.5000 } &  \num{ 0.5215 } &  \num{ 0.8019 } &  \num{ 0.1981 } \\ 

		\cline{1-9}
		
		\multirow{8}{*}{\rotatebox[origin=c]{90}{\footnotesize \text{\textbf{MrBanks 2}}}}

		& \multirow{8}{*}{\rotatebox[origin=c]{90}{\footnotesize \text{\textbf{ Ply, Sit (3) }}}}
		& SIMSBM(1,1)$^b$ &  \num{ 0.7032 +- 0.0001 } &  \num{ 0.6700 +- 0.0003 } &  \num{ 0.7049 +- 0.0002 } &  \num{ 0.7018 +- 0.0002 } &  \num{ 0.8350 +- 0.0002 } &  \num{ 0.1650 +- 0.0002 } \\ 
		& & SIMSBM(1,2)$^d$ &  \num{ 0.7032 +- 0.0002 } &  \num{ 0.6679 +- 0.0005 } &  \num{ 0.7028 +- 0.0004 } &  \num{ 0.7010 +- 0.0004 } &  \num{ 0.8340 +- 0.0003 } &  \num{ 0.1660 +- 0.0003 } \\ 
		& & SIMSBM(1,3) & \maxf{ \num{ 0.7290 +- 0.0003 } } & \maxf{ \num{ 0.7067 +- 0.0006 } } & \maxf{ \num{ 0.7547 +- 0.0005 } } & \maxf{ \num{ 0.7530 +- 0.0006 } } & \maxf{ \num{ 0.8533 +- 0.0003 } } & \maxf{ \num{ 0.1467 +- 0.0003 } } \\ 
		& & TF &  \num{ 0.6775 } &  \num{ 0.5953 } &  \num{ 0.5054 } &  \num{ 0.5259 } &  \num{ 0.7976 } &  \num{ 0.2024 } \\ 
		& & NMF &  \num{ 0.7137 } &  \num{ 0.6908 } &  \num{ 0.7246 } &  \num{ 0.7128 } &  \num{ 0.8397 } &  \num{ 0.1603 } \\ 
		& & KNN &  \num{ 0.7100 } &  \num{ 0.6699 } &  \num{ 0.7126 } &  \num{ 0.6856 } &  \num{ 0.8349 } &  \num{ 0.1651 } \\ 
		& & NB &  \num{ 0.6802 } &  \num{ 0.6512 } &  \num{ 0.6329 } &  \num{ 0.6225 } &  \num{ 0.8256 } &  \num{ 0.1744 } \\ 
		& & BL &  \num{ 0.6775 } &  \num{ 0.5953 } &  \num{ 0.5000 } &  \num{ 0.5181 } &  \num{ 0.7976 } &  \num{ 0.2024 } \\ 

		\cline{1-9}
		
		\multirow{8}{*}{\rotatebox[origin=c]{90}{\footnotesize \text{\textbf{Spotify}}}}

		& \multirow{8}{*}{\rotatebox[origin=c]{90}{\footnotesize \text{\textbf{ Artists (3)}}}}
		& SIMSBM(1)$^a$ &  \num{ 0.1741 +- 0.0004 } &  \num{ 0.2155 +- 0.0007 } & \maxf{ \num{ 0.7908 +- 0.0006 } } &  \num{ 0.1603 +- 0.0003 } &  \num{ 0.3827 +- 0.0004 } &  \num{ 0.0786 +- 0.0003 } \\ 
		& & SIMSBM(2)$^c$ &  \num{ 0.3156 +- 0.0005 } & \maxf{ \num{ 0.3348 +- 0.0004 } } &  \num{ 0.7661 +- 0.0005 } &  \num{ 0.2545 +- 0.0003 } & \maxf{ \num{ 0.4528 +- 0.0003 } } &  \num{ 0.0938 +- 0.0006 } \\ 
		& & SIMSBM(3) & \maxf{ \num{ 0.3243 +- 0.0004 } } &  \num{ 0.3209 +- 0.0003 } &  \num{ 0.7384 +- 0.0006 } & \maxf{ \num{ 0.2613 +- 0.0003 } } &  \num{ 0.4366 +- 0.0003 } &  \num{ 0.1079 +- 0.0007 } \\ 
		& & TF &  \num{ 0.0262 } &  \num{ 0.0042 } &  \num{ 0.4805 } &  \num{ 0.0159 } &  \num{ 0.0962 } &  \num{ 0.1550 } \\ 
		& & NMF &  \num{ 0.0371 } &  \num{ 0.0658 } &  \num{ 0.5650 } &  \num{ 0.0403 } &  \num{ 0.1762 } &  \num{ 0.2557 } \\ 
		& & KNN &  \num{ 0.3201 } &  \num{ 0.3009 } &  \num{ 0.7079 } &  \num{ 0.2400 } &  \num{ 0.3941 } &  \num{ 0.5212 } \\ 
		& & NB &  \num{ 0.0463 } &  \num{ 0.0846 } &  \num{ 0.7005 } &  \num{ 0.0576 } &  \num{ 0.2264 } & \maxf{ \num{ 0.0763 } } \\ 
		& & BL &  \num{ 0.0262 } &  \num{ 0.0532 } &  \num{ 0.5000 } &  \num{ 0.0135 } &  \num{ 0.1879 } &  \num{ 0.0969 } \\ 

		\cline{1-9}
		
		\multirow{8}{*}{\rotatebox[origin=c]{90}{\footnotesize \text{\textbf{PubMed}}}}

		& \multirow{8}{*}{\rotatebox[origin=c]{90}{\footnotesize \text{\textbf{ Symptoms (3)}}}}
		& SIMSBM(1)$^a$ &  \num{ 0.2915 +- 0.0002 } &  \num{ 0.5576 +- 0.0004 } &  \num{ 0.7475 +- 0.0001 } &  \num{ 0.2658 +- 0.0001 } &  \num{ 0.4641 +- 0.0001 } &  \num{ 0.2033 +- 0.0001 } \\ 
		& & SIMSBM(2)$^c$ &  \num{ 0.3127 +- 0.0001 } &  \num{ 0.5704 +- 0.0001 } &  \num{ 0.7613 +- 0.0001 } &  \num{ 0.2840 +- 0.0001 } &  \num{ 0.4838 +- 0.0001 } &  \num{ 0.1991 +- 0.0001 } \\ 
		& & SIMSBM(3) & \maxf{ \num{ 0.3219 +- 0.0001 } } & \maxf{ \num{ 0.5790 +- 0.0001 } } & \maxf{ \num{ 0.7666 +- 0.0001 } } & \maxf{ \num{ 0.2895 +- 0.0001 } } & \maxf{ \num{ 0.4937 +- 0.0001 } } & \maxf{ \num{ 0.1983 +- 0.0001 } } \\ 
		& & TF &  \num{ 0.1607 } &  \num{ 0.1003 } &  \num{ 0.5605 } &  \num{ 0.1777 } &  \num{ 0.1370 } &  \num{ 0.5118 } \\ 
		& & NMF &  \num{ 0.1606 } &  \num{ 0.0293 } &  \num{ 0.5368 } &  \num{ 0.2158 } &  \num{ 0.2321 } &  \num{ 0.2959 } \\ 
		& & KNN &  \num{ 0.2414 } &  \num{ 0.3251 } &  \num{ 0.6154 } &  \num{ 0.2324 } &  \num{ 0.2891 } &  \num{ 0.7730 } \\ 
		& & NB &  \num{ 0.2600 } &  \num{ 0.1618 } &  \num{ 0.7054 } &  \num{ 0.2389 } &  \num{ 0.2036 } &  \num{ 0.3058 } \\ 
		& & BL &  \num{ 0.1607 } &  \num{ 0.1003 } &  \num{ 0.5000 } &  \num{ 0.1026 } &  \num{ 0.2464 } &  \num{ 0.2834 } \\ 

		\cline{1-9}
		
		\multirow{7}{*}{\rotatebox[origin=c]{90}{\footnotesize \text{\textbf{Imdb 1}}}}

		& \multirow{7}{*}{\rotatebox[origin=c]{90}{\footnotesize \text{\textbf{ Usr, Cast (2) }}}}
		& SIMSBM(1,1)$^b$ & \maxf{ \num{ 0.3212 +- 0.0001 } } & \maxf{ \num{ 0.2434 +- 0.0001 } } & \maxf{ \num{ 0.6265 +- 0.0001 } } & \maxf{ \num{ 0.2502 +- 0.0001 } } &  \num{ 0.4360 +- 0.0001 } &  \num{ 0.3504 +- 0.0001 } \\ 
		& & SIMSBM(1,2)$^d$ & \num{ 0.2546 +- 0.0001 } &  \num{ 0.1006 +- 0.0038 } &  \num{ 0.4998 +- 0.0003 } &  \num{ 0.1509 +- 0.0001 } &  \num{ 0.2928 +- 0.0042 } &  \num{ 0.4527 +- 0.0051 } \\ 
		& & TF &  \num{ 0.2546 } &  \num{ 0.2300 } &  \num{ 0.4960 } &  \num{ 0.1485 } &  \num{ 0.4568 } &  \num{ 0.2702 } \\ 
		& & NMF &  \num{ 0.1329 } &  \num{ 0.0593 } &  \num{ 0.5007 } &  \num{ 0.1531 } &  \num{ 0.1549 } &  \num{ 0.8087 } \\ 
		& & KNN &  \num{ 0.2578 } &  \num{ 0.1899 } &  \num{ 0.5489 } &  \num{ 0.1679 } &  \num{ 0.3290 } &  \num{ 0.5328 } \\ 
		& & NB &  \num{ 0.2555 } &  \num{ 0.2351 } &  \num{ 0.5308 } &  \num{ 0.1607 } &  \maxf{ \num{ 0.4619 } } &  \maxf{ \num{ 0.2596 } } \\ 
		& & BL &  \num{ 0.2546 } &  \num{ 0.2300 } &  \num{ 0.5000 } &  \num{ 0.1508 } &  \num{ 0.4605 } &  \num{ 0.2586 } \\ 
		
		\cline{1-9}

		\multirow{6}{*}{\rotatebox[origin=c]{90}{\footnotesize \text{\textbf{Imdb 2}}}}

		& \multirow{6}{*}{\rotatebox[origin=c]{90}{\footnotesize \text{\textbf{Usr, Dir, Cast}}}}
		& SIMSBM(1,1,1) & \maxf{ \num{ 0.3896 +- 0.0001 } } & \maxf{ \num{ 0.3437 +- 0.0002 } } & \maxf{ \num{ 0.7593 +- 0.0001 } } & \maxf{ \num{ 0.3293 +- 0.0002 } } & \maxf{ \num{ 0.5705 +- 0.0001 } } & \maxf{ \num{ 0.1654 +- 0.0001 } } \\ 
		& & TF &  \num{ 0.2547 } &  \num{ 0.2238 } &  \num{ 0.5039 } &  \num{ 0.1513 } &  \num{ 0.4549 } &  \num{ 0.2636 } \\ 
		& & NMF &  \num{ 0.1127 } &  \num{ 0.0483 } &  \num{ 0.5005 } &  \num{ 0.1529 } &  \num{ 0.1406 } &  \num{ 0.8319 } \\ 
		& & KNN &  \num{ 0.2596 } &  \num{ 0.1890 } &  \num{ 0.5501 } &  \num{ 0.1681 } &  \num{ 0.3268 } &  \num{ 0.5248 } \\ 
		& & NB &  \num{ 0.2558 } &  \num{ 0.2373 } &  \num{ 0.5362 } &  \num{ 0.1617 } &  \num{ 0.4632 } &  \num{ 0.2571 } \\ 
		& & BL &  \num{ 0.2547 } &  \num{ 0.2286 } &  \num{ 0.5000 } &  \num{ 0.1507 } &  \num{ 0.4598 } &  \num{ 0.2574 } \\

		\cline{1-9}
	\end{tabular}
	}}
\end{table*}

\subsection{Datasets and evaluation protocol}
\subsubsection{Datasets}
The datasets we consider here are presented in Table~\ref{table-DS}. Each of them is made available along with the implementation of our model and our experiments on GitHub\footnote{Datasets and implementation available on \url{https://github.com/GaelPouxMedard/SIMSBM}}.

The \textbf{MrBanks datasets} has been gathered from a social experiment detailed in \cite{Guttieres2016MrBanks}. The experiment takes the form of a game where a player must guess whether a stock market curve will go up or down at the next time step. In order to do so, she can access various pieces of information, from which we selected the most relevant subset based on the description in \cite{Guttieres2016MrBanks,Poux2021MMSBMMrBanks}: direction of the market on the previous day (up/down), whether she guessed right (yes/no), and an expert's advice who is correct 60\% of the time (up/down/not consulted). Those are the 7 interacting pieces of information that define a situation. If the model considers pair interactions for instance, a situation can be defined as ``market went down and user guessed wrong'', or ``market went up and expert advised up''. A triplet interaction allows to get the full picture according to the selected pieces of information. In addition, we have access to the players age and gender. The goal is to predict whether the user will guess up or down given the available information.

For the \textbf{Spotify dataset} we collected user-made playlists on Spotify using the Spotipy python API. Our goal is to predict which next artist the user will add to the playlist, given the previous artists he already added. We consider the last 4 artists added by the user and their interaction to guess the next one. Note that it often happens for an artist to be added several times in a row.

The \textbf{PubMed dataset} is made of medical reports we gathered using the PubMed API. We use provided keywords to isolate symptoms and diseases in the text, as in \cite{Zhou2014SymptDisPubMed}. Our goal is to guess which diseases are discussed in the article given the symptoms that are listed in the document. Our guess is that a combination of symptoms helps narrowing the set of possible diagnoses. 

Finally, the \textbf{Imdb datasets} are provided and discussed in \cite{Harper2015ImdbDataset}. The original dataset comes with information about movies such as the lead actors starring in it and the movie's director. It also provides a list of users' ratings on movies. We aim to predict which rating a movie will get according to several combinations of parameters and their interactions: who directed the movie, who played in it, who gave the rating, etc.

90\% of each dataset's documents are used as a training set, and the other 10\% are used as an evaluation set. Each iteration of the SIMSBM is run 100 times on every dataset. The EM algorithm stops once the relative variation of the likelihood falls below $10^{-4}$ for 30 iterations in a row. We present the average results over all the runs. The number of clusters has been chosen based on the existing literature on similar datasets (Imdb \cite{Antonia2016AccurateAndScalableRS}, MrBanks \cite{Poux2021MMSBMMrBanks}, Spotify and PubMed \cite{Poux2021IMMSBM}) for demonstration purposes; dedicated work would be needed to infer their optimal number for every dataset.

Finally, when SIMSBM is evaluated on a dataset containing more interactions than it is designed to consider, the model is trained on the lower-order corresponding dataset. For instance, imagine a dataset considering one type interacting three times. This dataset is made of one observation only $(1, 2, 3, o)$. A SIMSBM iteration that considers pair interactions will then be trained on triplets $(1,2,o), (1,3,o)$ and $(2,3,o)$, and evaluation will be performed accordingly.

\subsubsection{Baselines and evaluation}
Evaluation is done according to the maximum F1 score, the precision at 1 (P@1), the area under the ROC curve, the area under the Precision-Recall curve (or Average Precision); since the problem is about multi-label classification, we consider the weighted version of these metrics --metrics are computed individually for each class, and averaged with each classes' weight being equal to the number of true instances in the class. The presented results are averaged over all 100 runs. We also consider the rank average precision and the normalized covering error (only here lower is better).\footnote{\label{note1}We used the sklearn Python library implementation.}. These last two metrics account for label ranking performance.
We compare to several standard baselines:
\begin{itemize}
    \item BL: the most naive baseline, where each output is predicted according to its frequency in the training set, without any context.
    \item NB$^{\text{\ref{note1}}}$: the Naive Bayes baseline assumes conditional independence between the input entities and updates the posterior probability according to Bayes law.
    \item KNN$^{\text{\ref{note1}}}$: K-nearest-neighbors. The output probabilities for a given entities array are defined according to a majority vote among the most similar entities arrays.
    \item NMF$^{\text{\ref{note1}}}$ and TF: Tensor Factorization baselines. For TF, we use the implementation provided by authors of \cite{Karatzoglou2010TF}. As discussed in the introduction, to make these methods fit to our problem, we have to define a continuous quantity to train the model. Instead of requiring an additional model to map possible outputs into a continuous space, we train the model on the frequency of outputs in a given context. Since NMF can only consider one entity as an input, we consider each different context as an independent entity. Outputs are added as an additional dimension to the data matrix instead of being a proper objective --their frequency is now the objective. The TF baseline is run for the same number of clusters as for the SIMSBM.
    \item MMSBM \cite{Airoldi2008MMSBM}, Bipartite-MMSBM \cite{Antonia2016AccurateAndScalableRS}, IMMSBM \cite{Poux2021IMMSBM}, T-MBM \cite{Tarres2019TMBM}: as discussed before, each of these models are special cases of SIMSBM. For presentation purpose, for each model, we keep the SIMSBM notation and indicate in superscript which of these it reduces to in this context. MMSBM=$^{\text{a}}$; Bi-MMSBM=$^{\text{b}}$; IMMSBM=$^{\text{c}}$; T-MBM=$^{\text{d}}$.
\end{itemize}

\subsection{Discussion}
\subsubsection{Main results}
We present our main results in Table~\ref{table-res}. In this table, we see that our formulation systematically outperforms the proposed baselines, as well as the ones it generalizes. In most cases, the possibility to add a layer or to consider higher-order interactions improves the performance over the existing baselines (MMSBM, Bi-MMSBM, IMMSBM and T-MBM). About the Spotify dataset, as stated before, artists are often added to a playlist in a row, leading to the probability of the next artist being the same as the one immediately before her to be higher. In this context, adding interaction terms adds noise in the modeling. This explains why the triple interactions version of SIMSBM does not perform better than its pair-interactions \cite{Antonia2016AccurateAndScalableRS} or no-interaction \cite{Airoldi2008MMSBM} iterations.

Besides numerical results, Table~\ref{table-res} underlines how easy the SIMSBM makes it to design tailored models for a variety of different problems. Using a single framework, we could consider input context sizes ranging from 1 to 4 entries, and interaction orders going from 1 to 3. Model selection (or in this case, SIMSBM iteration selection) becomes simpler as a result, since different input settings can be readily tested. Up to now, adding a new layer (e.g. T-MBM with respect to Bi-MMSBM) or increasing the interaction order (e.g. IMMSBM w.r.t. MMSBM) required a new publication to explain the advance; SIMSBM unifies these approaches into one global model, each iteration of which is readily usable in a number applications, as demonstrated in Table~\ref{table-res}.

In a similar line of reasoning, we see that SIMSBM is finds applications in the field of recommender systems. This is expected, as it generalizes two works that have been developed for explicit recommendation problems \cite{Antonia2016AccurateAndScalableRS,Poux2021IMMSBM}. We emphasize this point running SIMSBM on recommendation datasets such as Imdb and Spotify --and performing replication studies, see next section. However, the field of application of SIMSBM ranges beyond the sole recommender systems aspect, as shown tackling an assisted diagnosis problem (PubMed dataset) and choice mechanisms data-driven study (MrBanks dataset, \cite{Guttieres2016MrBanks,Poux2021MMSBMMrBanks}). 

A final note on the running time of the algorithm is necessary. We saw with Eq.~\ref{eq:maxTheta} that one iteration of the EM algorithm can be performed with complexity $\mathcal{O}(\vert R^{\circ} \vert)$. However, the combinatorial aspect of considering larger input sizes makes $\vert R^{\circ} \vert$ grow exponentially with the context size and interaction order. This aspect must be considered when running large scale experiments. Typically, in the case of the Spotify dataset, 200 input context entries (artists) were considered. When considering one order of interactions, $\vert R^{\circ} \vert$ scales as a fraction $x$ of 200. When considering two orders of interaction, $\vert R^{\circ} \vert$ scales as the same fraction $x$ of 200$\times$200; when considering three orders of interaction, $\vert R^{\circ} \vert$ scales as the same fraction $x$ of 200$\times$200$\times$200; etc. This effect is less crucial when considering larger context sizes, because the number of new n-plets in $R^{\circ}$ does not necessarily scale as the product of input entities type's size. For instance in Imdb 2, a movie director tends to record movies using a reduced set of lead actors; the number of different entries $R^{\circ}$ does not grow much in this case. The main point regarding complexity, is that iterations of the EM algorithm scale with the size of the dataset $\vert R^{\circ} \vert$; it is however left to the SIMSBM user to consider that $\vert R^{\circ} \vert$ can grow large depending on the task at hand.

\begin{table*}
	\centering    
	\caption{Replication results on two datasets used in (Godoy-Lorite et al., 2016a) and (Poux-Medard et al., 2021a), referenced in the main text. The standard error on the last digits over all 100 runs is indicated between parenthesis. Overall, we retrieve the same results as those presented in (Godoy-Lorite et al., 2016a) and (Poux-Medard et al., 2021a).}
    \label{table-rep}
    \noindent\makebox[\textwidth]{\resizebox{\textwidth}{!}{
	\begin{tabular}{|l|l|l|S|S|S|S|S|S|S}

		\cline{1-9}
		& & & \text{F1} & \text{P@1} & \text{AUCROC} & \text{AUCPR} & \text{RankAvgPrec} & \text{CovErrNorm} \\ 

		\cline{1-9}
		\multirow{5}{*}{\rotatebox[origin=c]{90}{\scriptsize \text{\textbf{Imdb \cite{Antonia2016AccurateAndScalableRS}}}}}

		& \multirow{5}{*}{\rotatebox[origin=c]{90}{\footnotesize \text{\textbf{User, Movie}}}}
		& SIMSBM(1,1) & \maxf{ \num{ 0.3995 +- 0.0002 } } & \maxf{ \num{ 0.3558 +- 0.0003 } } & \maxf{ \num{ 0.7665 +- 0.0001 } } & \maxf{ \num{ 0.3406 +- 0.0003 } } & \maxf{ \num{ 0.5805 +- 0.0002 } } & \maxf{ \num{ 0.1593 +- 0.0001 } } \\ 
		& & TF &  \num{ 0.2570 } &  \num{ 0.2348 } &  \num{ 0.5031 } &  \num{ 0.1541 } &  \num{ 0.4627 } &  \num{ 0.2573 } \\ 
		& & KNN &  \num{ 0.2668 } &  \num{ 0.2002 } &  \num{ 0.5558 } &  \num{ 0.1735 } &  \num{ 0.3308 } &  \num{ 0.4834 } \\ 
		& & NB &  \num{ 0.2585 } &  \num{ 0.2382 } &  \num{ 0.5377 } &  \num{ 0.1660 } &  \num{ 0.4664 } &  \num{ 0.2536 } \\ 
		& & BL &  \num{ 0.2570 } &  \num{ 0.2349 } &  \num{ 0.5000 } &  \num{ 0.1525 } &  \num{ 0.4647 } &  \num{ 0.2557 } \\ 

		\cline{1-9}
		
		\multirow{5}{*}{\rotatebox[origin=c]{90}{\scriptsize \text{\textbf{MrBanks \cite{Guttieres2016MrBanks}}}}}

		& \multirow{5}{*}{\rotatebox[origin=c]{90}{\footnotesize \text{\textbf{Ply, Full sit}}}}
		& SIMSBM(1,1) & \maxf{ \num{ 0.7126 +- 0.0002 } } & \maxf{ \num{ 0.6688 +- 0.0004 } } & \maxf{ \num{ 0.7126 +- 0.0003 } } & \maxf{ \num{ 0.7180 +- 0.0004 } } & \maxf{ \num{ 0.8344 +- 0.0002 } } & \maxf{ \num{ 0.1656 +- 0.0002 } } \\ 
		& & TF &  \num{ 0.6795 } &  \num{ 0.6037 } &  \num{ 0.5176 } &  \num{ 0.5363 } &  \num{ 0.8019 } &  \num{ 0.1981 } \\ 
		& & KNN &  \num{ 0.6940 } &  \num{ 0.6433 } &  \num{ 0.6668 } &  \num{ 0.6430 } &  \num{ 0.8217 } &  \num{ 0.1783 } \\ 
		& & NB &  \num{ 0.6795 } &  \num{ 0.6037 } &  \num{ 0.5907 } &  \num{ 0.5822 } &  \num{ 0.8019 } &  \num{ 0.1981 } \\ 
		& & BL &  \num{ 0.6795 } &  \num{ 0.6037 } &  \num{ 0.5000 } &  \num{ 0.5215 } &  \num{ 0.8019 } &  \num{ 0.1981 } \\ 

		\cline{1-9}
	\end{tabular}
	}}
\end{table*}
\subsubsection{Replication studies}
To further underline the applicability domain of SIMSBM, we ran it on the datasets considered in \cite{Antonia2016AccurateAndScalableRS} and in \cite{Poux2021MMSBMMrBanks}. We chose the parameters so that SIMSBM runs using the same model's specifications. Our results are shown in Table~\ref{table-rep}. They are similar as those of \cite{Antonia2016AccurateAndScalableRS,Poux2021MMSBMMrBanks}. It confirms that our model correctly generalizes existing models tackling similar problems.

Interestingly in the case of \cite{Poux2021MMSBMMrBanks}, the authors propose to describe a given situation in form of a unique key, where each key is independent from the others (Table~\ref{table-rep}, MrBanks). Our formulation with triple interactions (Table~\ref{table-res}, MrBanks 2) improves the results on the same dataset the authors provided. This is because the constituents of a situation are not independent anymore but instead behave as elementary interacting pieces of context, which provides a more accurate description of reality: a situation is not considered as a whole anymore, but instead as the combination of several pieces of information.

\section{Conclusion}
In this paper, we developed a global framework, SIMSBM, that generalizes several existing models from the literature. We derived an expectation-maximization algorithm that runs in linear time with the number of observations. We then demonstrated that SIMSBM recover several models from the literature as special cases, such as MMSBM, Bi-MMSBM, IMMSBM and T-MBM. 
This results in a highly flexible model that can be applied to a broad range of classification problems, as shown through systematic evaluation of the proposed formulation on several real-world datasets. In particular, we cited throughout the text a number of experimental studies conducted in medicine, social behaviour and recommendation using special cases of our model; using alternative iterations of the SIMSBM framework may help further improve the description and understanding of the interacting processes at stake between an arbitrary greater number of input entities.

\bibliographystyle{IEEEtran}
\bibliography{Bibliography.bib}

\begin{thebibliography}{10}
\providecommand{\url}[1]{#1}
\csname url@samestyle\endcsname
\providecommand{\newblock}{\relax}
\providecommand{\bibinfo}[2]{#2}
\providecommand{\BIBentrySTDinterwordspacing}{\spaceskip=0pt\relax}
\providecommand{\BIBentryALTinterwordstretchfactor}{4}
\providecommand{\BIBentryALTinterwordspacing}{\spaceskip=\fontdimen2\font plus
\BIBentryALTinterwordstretchfactor\fontdimen3\font minus
  \fontdimen4\font\relax}
\providecommand{\BIBforeignlanguage}[2]{{%
\expandafter\ifx\csname l@#1\endcsname\relax
\typeout{** WARNING: IEEEtran.bst: No hyphenation pattern has been}%
\typeout{** loaded for the language `#1'. Using the pattern for}%
\typeout{** the default language instead.}%
\else
\language=\csname l@#1\endcsname
\fi
#2}}
\providecommand{\BIBdecl}{\relax}
\BIBdecl

\bibitem{Koren2009MFReco}
Y.~Koren, R.~Bell, and C.~Volinsky, ``Matrix factorization techniques for
  recommender systems,'' \emph{Computer}, vol.~42, no.~8, pp. 30--37, 2009.

\bibitem{Fund2007OnlineNMF1}
Y.-H. Fung, C.-H. Li, and W.~K. Cheung, ``Online discussion participation
  prediction using non-negative matrix factorization,'' in \emph{Proceedings of
  the 2007 IEEE/WIC/ACM International Conferences on Web Intelligence and
  Intelligent Agent Technology - Workshops}, ser. WI-IATW '07.\hskip 1em plus
  0.5em minus 0.4em\relax USA: IEEE Computer Society, 2007, p. 284–287.

\bibitem{Cao2007OnlineNMF2}
B.~Cao, D.~Shen, J.-T. Sun, X.~Wang, Q.~Yang, and Z.~Chen, ``Detect and track
  latent factors with online nonnegative matrix factorization,'' in
  \emph{IJCAI}, 2007.

\bibitem{Karatzoglou2010TF}
\BIBentryALTinterwordspacing
A.~Karatzoglou, X.~Amatriain, L.~Baltrunas, and N.~Oliver, ``Multiverse
  recommendation: N-dimensional tensor factorization for context-aware
  collaborative filtering,'' in \emph{Proceedings of the Fourth ACM Conference
  on Recommender Systems}, ser. RecSys '10.\hskip 1em plus 0.5em minus
  0.4em\relax New York, NY, USA: Association for Computing Machinery, 2010, p.
  79–86. [Online]. Available: \url{https://doi.org/10.1145/1864708.1864727}
\BIBentrySTDinterwordspacing

\bibitem{Hidasi2012Itals}
B.~Hidasi and D.~Tikk, ``Fast als-based tensor factorization for context-aware
  recommendation from implicit feedback,'' in \emph{Machine Learning and
  Knowledge Discovery in Databases}.\hskip 1em plus 0.5em minus 0.4em\relax
  Springer Berlin Heidelberg, 2012, pp. 67--82.

\bibitem{Bhargava2015}
\BIBentryALTinterwordspacing
P.~Bhargava, T.~Phan, J.~Zhou, and J.~Lee, ``Who, what, when, and where:
  Multi-dimensional collaborative recommendations using tensor factorization on
  sparse user-generated data,'' in \emph{Proceedings of the 24th International
  Conference on World Wide Web}, ser. WWW '15.\hskip 1em plus 0.5em minus
  0.4em\relax Republic and Canton of Geneva, CHE: International World Wide Web
  Conferences Steering Committee, 2015, p. 130–140. [Online]. Available:
  \url{https://doi.org/10.1145/2736277.2741077}
\BIBentrySTDinterwordspacing

\bibitem{Harshman1970CP}
R.~A. Harshman, ``Foundations of the parafac procedure: Models and conditions
  for an “explanatory” multi-modal factor analysis,'' \emph{UCLA Working
  Papers in Phonetics}, vol.~16, p. 1–84, 1970.

\bibitem{Carroll1970CP}
J.~Carroll and J.~Chang, ``Analysis of individual differences in
  multidimensional scaling via an n-way generalization of “eckart-young”
  decomposition,'' \emph{Psychometrika}, vol.~35, pp. 283--319, 1970.

\bibitem{Acar2010CP1}
E.~Acar, D.~M. Dunlavy, T.~G. Kolda, and M.~M{\o}rup, ``Scalable tensor
  factorizations with missing data,'' in \emph{SDM10: Proceedings of the 2010
  SIAM International Conference on Data Mining}, 2010, pp. 701--712.

\bibitem{Filipovi2015CP2}
\BIBentryALTinterwordspacing
M.~Filipovi\'{c} and A.~Juki\'{c}, ``Tucker factorization with missing data
  with application to low-nn-rank tensor completion,'' \emph{Multidimensional
  Syst. Signal Process.}, vol.~26, no.~3, p. 677–692, jul 2015. [Online].
  Available: \url{https://doi.org/10.1007/s11045-013-0269-9}
\BIBentrySTDinterwordspacing

\bibitem{Antonia2016AccurateAndScalableRS}
A.~Godoy-Lorite, R.~Guimerà, C.~Moore, and M.~Sales-Pardo, ``Accurate and
  scalable social recommendation using mixed-membership stochastic block
  models,'' \emph{PNAS}, vol. 113, no.~50, pp. 14\,207--14\,212, 2016.

\bibitem{Tarres2019TMBM}
M.~Tarr\'es-Deulofeu, A.~Godoy-Lorite, R.~Guimer\`a, and M.~Sales-Pardo,
  ``Tensorial and bipartite block models for link prediction in layered
  networks and temporal networks,'' \emph{Phys. Rev. E}, vol.~99, p. 032307,
  Mar 2019.

\bibitem{Poux2021IMMSBM}
\BIBentryALTinterwordspacing
G.~Poux-M\'edard, J.~Velcin, and S.~Loudcher, ``Interactions in information
  spread: quantification and interpretation using stochastic block models,''
  \emph{RecSys'21}, 2021. [Online]. Available:
  \url{doi.org/10.1145/3460231.3474254}
\BIBentrySTDinterwordspacing

\bibitem{Poux2022InterInfSpreadTheWebConf}
G.~Poux-Médard, ``Interactions in information spread,'' \emph{WWW '22:
  Companion Proceedings of the Web Conference 2022}, pp. 313--317, 2022.

\bibitem{Neal1998ConvergenceEM}
\BIBentryALTinterwordspacing
R.~M. Neal and G.~E. Hinton, \emph{A View of the Em Algorithm that Justifies
  Incremental, Sparse, and other Variants}.\hskip 1em plus 0.5em minus
  0.4em\relax Dordrecht: Springer Netherlands, 1998, pp. 355--368. [Online].
  Available: \url{https://doi.org/10.1007/978-94-011-5014-9_12}
\BIBentrySTDinterwordspacing

\bibitem{Holland1983SBM}
\BIBentryALTinterwordspacing
P.~W. Holland, K.~B. Laskey, and S.~Leinhardt, ``Stochastic blockmodels: First
  steps,'' \emph{Social Networks}, vol.~5, no.~2, pp. 109--137, 1983. [Online].
  Available:
  \url{https://www.sciencedirect.com/science/article/pii/0378873383900217}
\BIBentrySTDinterwordspacing

\bibitem{Guimera2012HumanPrefSBM}
R.~Guimera, A.~Llorente, and M.~Sales-Pardo, ``Predicting human preferences
  using the block structure of complex social networks,'' \emph{PLOS One},
  vol.~7, no.~9, 2012.

\bibitem{Funka2019ReviewSBM}
T.~Funke and T.~Becker, ``Stochastic block models: A comparison of variants and
  inference methods,'' \emph{PloS one}, 2019.

\bibitem{Airoldi2008MMSBM}
E.~Airoldi, D.~Blei, S.~Fienberg, and E.~Xing, ``Mixed membership stochastic
  blockmodels,'' \emph{Journal of Machine Learning Research}, vol.~9, pp.
  1991--1992, 2008.

\bibitem{Guimera2013DrugdrugSBM}
R.~Guimerà and M.~Sales-Pardo, ``A network inference method for large-scale
  unsupervised identification of novel drug-drug interactions,'' \emph{PLoS
  Comput Biol}, 2013.

\bibitem{CoboLopez2018SocialDilemma}
D.~J. Cobo-López~S., Godoy-Lorite~A., ``Optimal prediction of decisions and
  model selection in social dilemmas using block models,'' \emph{EPJ Data Sci},
  vol. 7(48), 2018.

\bibitem{Christakopoulou2014HOSLIM}
\BIBentryALTinterwordspacing
E.~Christakopoulou and G.~Karypis, ``Hoslim: Higher-order sparse linear method
  for top-n recommender systems,'' \emph{PAKDD}, 2014. [Online]. Available:
  \url{10.1007/978-3-319-06605-9_4}
\BIBentrySTDinterwordspacing

\bibitem{Steck2021Interactions}
\BIBentryALTinterwordspacing
H.~Steck and D.~Liang, ``Negative interactions for improved collaborative
  filtering: Don’t go deeper, go higher,'' \emph{RecSys'21}, p. 34–43,
  2021. [Online]. Available: \url{10.1145/3460231.3474273}
\BIBentrySTDinterwordspacing

\bibitem{Myers2012CoC}
S.~A. Myers and J.~Leskovec, ``Clash of the contagions: Cooperation and
  competition in information diffusion,'' \emph{2012 IEEE 12th International
  Conference on Data Mining}, pp. 539--548, 2012.

\bibitem{Poux2021MMSBMMrBanks}
G.~Poux-M\'edard, S.~Cobo-Lopez, J.~Duch, R.~Guimerà, and M.~Sales-Pardo,
  ``Complex decision-making strategies in a stock market experiment explained
  as the combination of few simple strategies,'' \emph{EPJ Data Science},
  vol.~10, 2021.

\bibitem{Yuchung1987}
Y.~J. W.~Y. Wong, ``Stochastic blockmodels for directed graphs,'' \emph{Journal
  of the American Statistical Association}, vol.~82, no. 397, pp. 8--19, 1987.

\bibitem{Godoy2016EmailNetwork}
\BIBentryALTinterwordspacing
A.~Godoy-Lorite, R.~Guimerà, and M.~Sales-Pardo, ``Long-term evolution of
  email networks: Statistical regularities, predictability and stability of
  social behaviors,'' \emph{PLOS ONE}, vol.~11, no.~1, pp. 1--11, 01 2016.
  [Online]. Available: \url{https://doi.org/10.1371/journal.pone.0146113}
\BIBentrySTDinterwordspacing

\bibitem{Guttieres2016MrBanks}
M.~Guti\'erres-Roig, C.~Segura, J.~Dutch, and J.~Perello, ``Market imitation
  and win-stay lose-shift strategies emerge as unintended patterns in market
  direction guesses,'' \emph{PLOS One}, 2016.

\bibitem{Zhou2014SymptDisPubMed}
X.~Zhou, J.~Menche, A.-L. Barabasi, and A.~Sharma, ``Human symptoms–disease
  network,'' \emph{Nature communications}, vol.~5, p. 4212, 2014.

\bibitem{Harper2015ImdbDataset}
\BIBentryALTinterwordspacing
F.~M. Harper and J.~A. Konstan, ``The movielens datasets: History and
  context,'' \emph{ACM Trans. Interact. Intell. Syst.}, vol.~5, no.~4, dec
  2015. [Online]. Available: \url{https://doi.org/10.1145/2827872}
\BIBentrySTDinterwordspacing

\end{thebibliography}

\end{document}